\documentclass[10pt,twocolumn,letterpaper]{article}

\usepackage[pagenumbers]{cvpr} % To force page numbers, e.g. for an arXiv version

\usepackage{graphicx}
\usepackage{amsmath}
\usepackage{amssymb}
\usepackage{booktabs}
\usepackage{multirow}
\usepackage[table,xcdraw]{xcolor}

\usepackage[pagebackref,breaklinks,colorlinks]{hyperref}

\usepackage[capitalize]{cleveref}
\crefname{section}{Sec.}{Secs.}
\Crefname{section}{Section}{Sections}
\Crefname{table}{Table}{Tables}
\crefname{table}{Tab.}{Tabs.}

\begin{document}

\title{MataDoc: Margin and Text Aware Document Dewarping for Arbitrary Boundary}

\author{Beiya Dai$^1$\thanks{Equal contribution.}, Xing li$^{2*}$, Qunyi Xie$^{2*}$, Yulin Li$^2$, Xiameng Qin$^2$, Chengquan Zhang$^2$, Kun Yao$^2$, Junyu Han$^2$ \\
$^1${National University of Defense Technology}, 
$^2${Baidu Inc.} \\
{\tt\small {daibeiya}@nudt.edu.cn, \{lixing14, xiequnyi, liyulin03\}@baidu.com} \\
{\tt\small \{qinxiameng, zhangchengquan, yaokun01, hanjunyu\}@baidu.com}
}

\maketitle

\begin{abstract}
Document dewarping from a distorted camera-captured image is of great value for OCR and document understanding. The document boundary plays an important role which is more evident than the inner region in document dewarping.
Current learning-based methods mainly focus on complete boundary cases, leading to poor document correction performance of documents with incomplete boundaries.
In contrast to these methods, this paper proposes \textbf{\textit{MataDoc}}, the first method focusing on arbitrary boundary document dewarping with margin and text aware regularizations.
Specifically, we design the margin regularization by explicitly considering background consistency to enhance boundary perception. Moreover, we introduce word position consistency to keep text lines straight in rectified document images. To produce a comprehensive evaluation of MataDoc, we propose a novel benchmark \textbf{ArbDoc}, mainly consisting of document images with arbitrary boundaries in four typical scenarios.
Extensive experiments confirm the superiority of MataDoc with consideration for the incomplete boundary on ArbDoc and also demonstrate the effectiveness of the proposed method on DocUNet, DIR300, and WarpDoc datasets.
\end{abstract}

%%%%%%%%% BODY TEXT
\section{Introduction}
\label{sec:intro}

\begin{figure}[t]
        \centering 
        \includegraphics[width=0.98\linewidth]{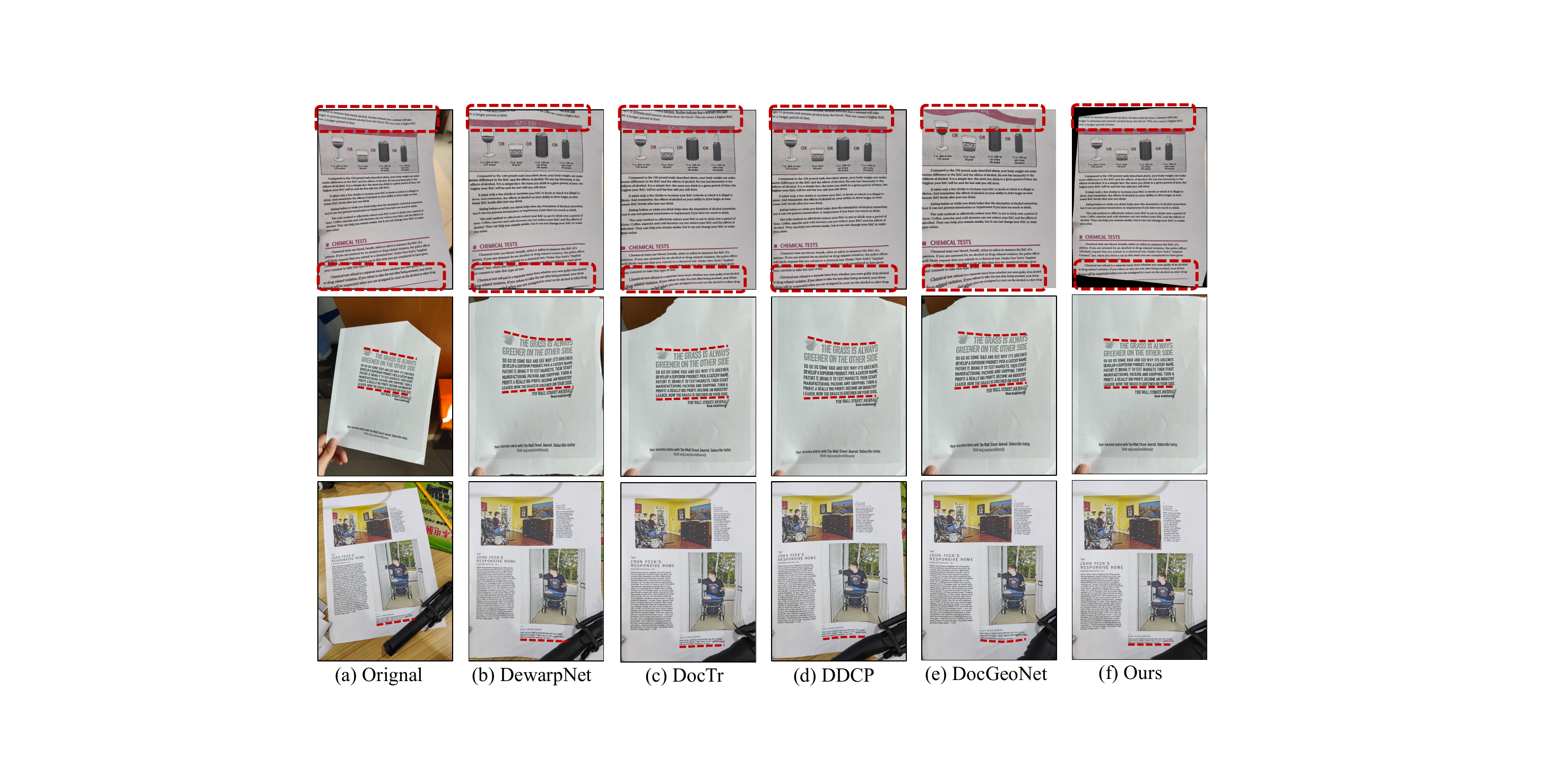}
        \caption{Four types of different document images (first row) from ArbDoc benchmark dataset. (a) Complete boundary. (b) Boundary overflow. (c) Corner absence. (d) Boundary occlusion. The images in the second row are the results of MataDoc.}
        \label{fig:1}
        \vspace{-1em}
    \end{figure}

    Document digitization using a handheld mobile device for transferring information from physical papers to electronic format has become very common in our daily lives. However, it usually leads to heavy geometrical deformation such as bending, folding, or creases in obtained document images. Besides, because of the relative position between the camera, physical document, and other objects in the environment, the captured document images may suffer from the incomplete boundary, as shown in Figure~\ref{fig:1}(b$\sim$d). Those problems may result in poor correction performance and harm downstream tasks, such as optical character recognition (OCR)~\cite{zhai2023textformer}, document understanding~\cite{li2021structext}, and even document readability~\cite{JiaxinZhang2022MariorMR}. 
    
\begin{figure*}[t]
        \centering 
        \includegraphics[width=0.98\linewidth]{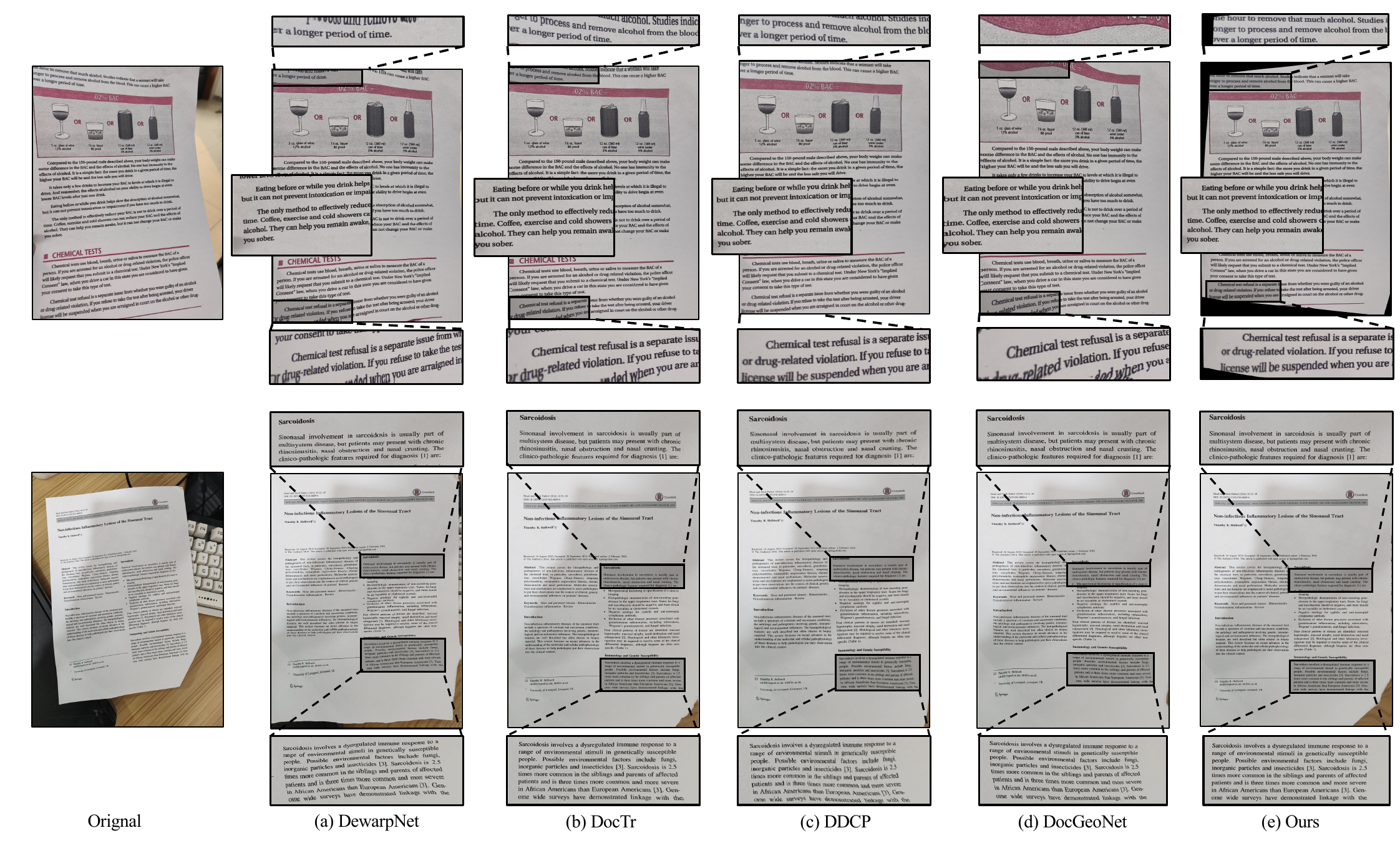}
        \caption{Qualitative comparison with previous methods on the ArbDoc benchmark dataset in terms of rectified images and local textline details. For the comparisons of rectified images, we crop some local patches and zoom in on local dewarping details. 
        }\label{fig:2}
        \vspace{-1em}
    \end{figure*}

    The incomplete boundary consists of three types: boundary overflow, corner absence, and boundary occlusion. Boundary overflow means that the document boundaries are out of the image, as shown in Figure~\ref{fig:1}(b). Corner absence (Figure~\ref{fig:1}(c)) happens when parts of the document boundary are over-folded, while boundary occlusion (Figure~\ref{fig:1}(d)) appears due to some foreground objects. Existing methods ~\cite{DasSagnik2019DewarpNetSD,HaoFeng2022DocTrDI,GuoWangXie2021DocumentDW,HaoFeng2022GeometricRL} mainly focus on improving the correction performance of document image with complete boundary, which can not handle arbitrary boundary situations. As shown in Figure~\ref{fig:2}(a$\sim$d), those methods can not correctly rectify the boundary areas while applying them to incomplete boundary situations. As a result, the rectified text lines near boundaries in horizontal and vertical directions are not straight enough, as shown in the first and second dewarping details, respectively. 

    In this paper, we propose \textbf{Ma}rgin and \textbf{t}ext \textbf{a}ware document dewarping network, called \textbf{\textit{MataDoc}}, the first method focusing on document dewarping with arbitrary boundary. To enhance boundary perception in incomplete situations, MataDoc proposes margin aware regularization. The Margin is defined as the background regions in the rectified mask~\cite{JiaxinZhang2022MariorMR}. Specifically, we use backward mapping to obtain the rectified mask from the distorted document mask. With the constraint of the marginal regions, boundaries can be explicitly defined to guide pixel-wise displacement learning.
    Moreover, we introduce the text aware regularization to keep rectified text lines straight. 
    We remap text lines in the rectified ground truth images to distorted form using the predicted backward map and then obtain the predicted rectified text lines with forward map ground truth, followed by constraining vertical coordinate consistency of the control points belonging to the same text line.
    To demonstrate the effectiveness of our approach, particularly on document images with incomplete boundaries, we present a challenging benchmark dataset with \textbf{Ar}bitrary \textbf{b}oundaries called ArbDoc. The dataset consists of 188 real document images with OCR annotations, which is divided into four different document boundary types of document images and covers rich document image application scenarios.

    \begin{figure*}
    \centering
    \includegraphics[width=0.85\linewidth]{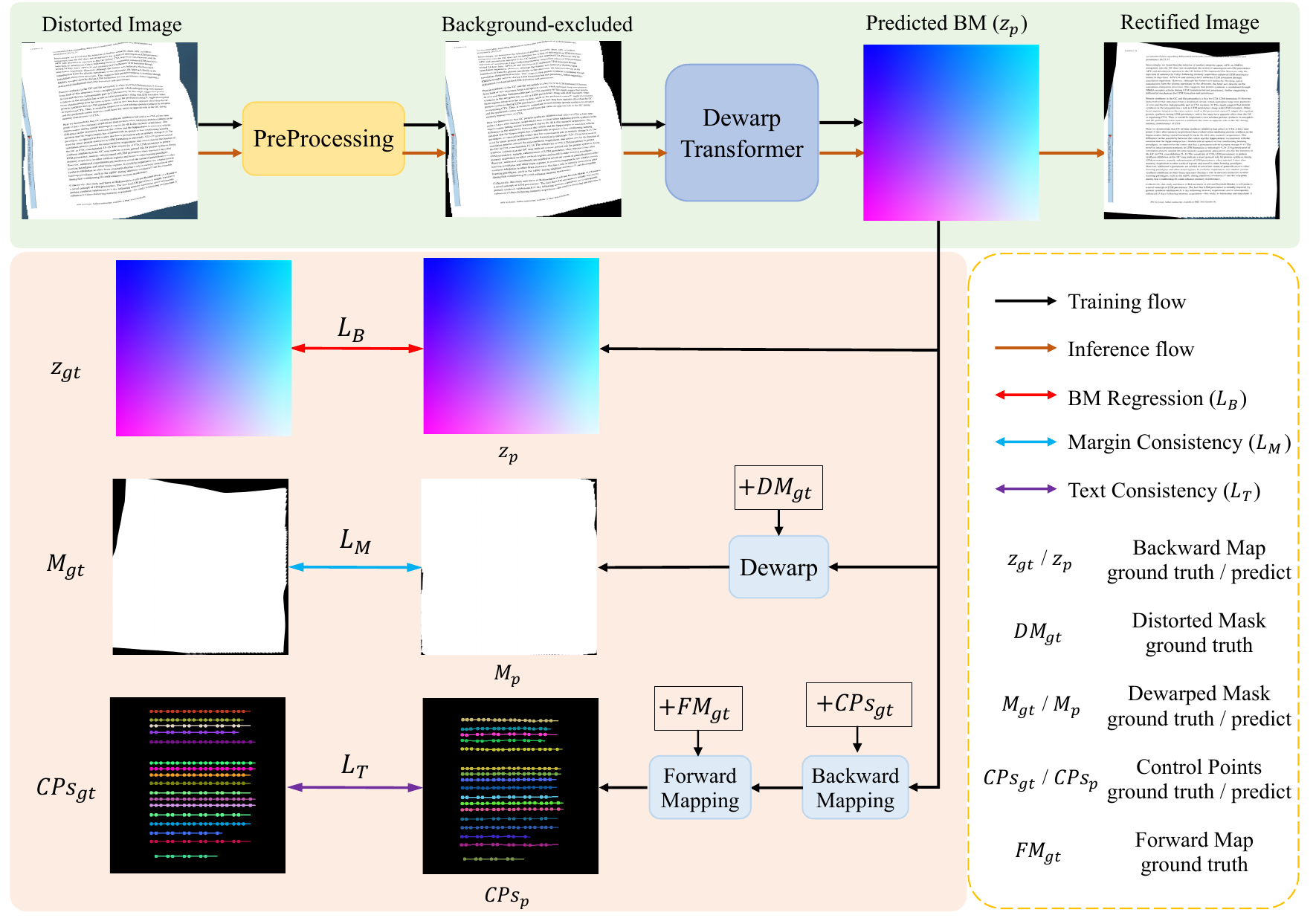}
    \caption{\textbf{The overall architecture of our proposed method MataDoc.}Taking distorted image as input, MataDoc removes the background firstly, then learns to predict backward map by \textit{Dewarp Transformer} and finally uses the predicted backward map for dewarping. MataDoc involves the backward map regression ($L_B$), text constraint ($L_T$) , and margin constrain ($L_M$) for dewarp training.}
    \label{fig:3}
    \end{figure*}

     The main contributions of this paper are summarized below:
    \begin{itemize}
    \item We propose the margin and text aware regularization (MataDoc) to rectify document images with arbitrary boundaries. To our knowledge, MataDoc is the first method to analyze and solve arbitrary boundary problems.
    
    \item In addition to the dewarping method, we present a new challenging benchmark called ArbDoc. The dataset comprises 188 real, OCR annotated, distorted document images with arbitrary boundaries.
    
    \item Extensive experiments demonstrate the effectiveness of MataDoc on ArbDoc, and the universality of MataDoc on DocUNet, DIR300, and WarpDoc datasets.
    
    \end{itemize}

\section{Related Work}

\subsection{Rectification based on deep learning}
    With the continuous development of deep learning and transformer~\cite{AshishVaswani2017AttentionIA} in computer vision~\cite{carion2020end,chen2021pre,dosovitskiy2020image}, document image dewarping model ~\cite{DasSagnik2019DewarpNetSD,XiyanLiu2020GeometricRO,GuoWangXie2021DocumentDW,feng2021docscanner,garai2021theoretical,garai2021dewarping,ramanna2019document,KeMa2022LearningFD} based on deep learning has become the mainstream.
    Compared with traditional methods~\cite{brown2001document,meng2014active,he2013book,cao2003rectifying}, methods based on deep learning have more robust feature expression capabilities. We classify deep learning-based document dewarping into forward mapping and backward mapping.
    
    \subsubsection{Rectification based on forward mapping}
        These methods predict a forward map whose value represents the position of rectified image that the corresponding pixel in the distorted image fills.
        DocUNet~\cite{KeMa2018DocUNetDI} predicts a forward mapping with a stacked U-Net and remaps each pixel on the distorted image to rectified one. 
        Patch-CNN~\cite{XiaoyuLi2019DocumentRA} seeks to learn the forward map on the patch level, followed by a robust stitching technique in the gradient domain.
        RDGR~\cite{XiangweiJiang2022RevisitingDI} obtains the forward graph by grid regularization and then converts the forward mapping into the backward mapping by LinearNDinterpolator.
        However, the forward mapping scheme usually requires complex post-processing operations to convert into backward mapping, which takes high computational cost.

    \subsubsection{Rectification based on backward mapping}
        Backward map based methods directly predict the coordinate mapping in the distorted image, whose value represents the coordinate of the sampled pixel in the distorted image. The rectified image can be obtained using grid sampling operation.
        Currently, most of the methods~\cite{AmirMarkovitz2020CanYR,DasSagnik2019DewarpNetSD,SagnikDas2021EndtoEndPU,HaoFeng2022DocTrDI,HaoFeng2022GeometricRL} directly predict the backward map and get rid of the post-processing step from the forward to the backward transformation.
        DewarpNet~\cite{DasSagnik2019DewarpNetSD} learns the correction model of documents by stacking 3D and 2D regression networks.
        DocTr~\cite{HaoFeng2022DocTrDI} learns the document correction model using a transformer based neural network.
        DocGeoNet~\cite{HaoFeng2022GeometricRL} leverages the U-Net model to learn the correction model for documents.
        In this work, we follow the backward map based approach due to its simplicity and make some optimizations to it.
        
    \subsection{Rectification Based on regular constraints}
    
        The regular constraint is an important and commonly used means of document image dewarping. Several methods employ curve-fitting loss to address methods based on B-spline curves~\cite{OlivierLavialle2001ActiveCN,GaofengMeng2012MetricRO}, nonlinear curves~\cite{ChanghuaWu2002DocumentID}, and polynomial approximations~\cite{LotharMischke2005DocumentID,BeomsuKim2015DocumentDV,TaehoKil2017RobustDI}. 
        Olivier et al.~\cite{OlivierLavialle2001ActiveCN} use cubic B-spline curves as geometric constraints to correct curved text lines. 
        Wu et al.~\cite{wu2002document} use nonlinear curves as geometric constraints to correct document images. 
        Mischke et al.~\cite{mischke2005document} uses a polynomial approximation to model curved text lines and perform document image correction. Some methods use content-aware loss to guide the network to pay more attention to content-rich regions during rectification. Marior~\cite{JiaxinZhang2022MariorMR} introduced a method based on a margin removal module and designed a novel content-aware loss function. Some methods use structural loss, which uses loss to supervise structural encoders for document geometry shaping. DocGeoNet~\cite{HaoFeng2022GeometricRL} uses a loss to supervise a structural encoder to learn 3D shapes, and a segmentation loss to guide a CNN-based text line extractor to learn text line features.
        In addition to text lines, document constraints such as boundaries~\cite{MichaelSBrown2006GeometricAS}, character~\cite{Zandifar2007UnwarpingSI}, line spacing, and text line direction~\cite{HyungIlKoo2010StateEI} can also be exploited.
        
        In contrast to the methods above, we incorporate text line and margin constraints to assist the model in improving the robustness of document dewarping with incomplete boundary problems.

\section{Methodology}

    As shown in Figure~\ref{fig:3}, given a distorted image, MataDoc removes the background of the document (section~\ref{section:3.1}) and then learns to predict the backward map by a Dewarp Transformer (section~\ref{section:3.2}). The predicted backward map is finally utilized for document dewarping (section~\ref{section:3.2}). Besides, we describe the details of proposed dewarp regularizations and the MataDoc training in section~\ref{Dewarp Regularizations}.
        
    \subsection{Preprocessing}
    \label{section:3.1}
    The preprocessing module includes two operations: object segmentation and textline extraction. In the background removal, object segmentation is used to locate the document and obtain the mask and image of the foreground. In particular, we utilize DeepLabv3+~\cite{LiangChiehChen2018EncoderDecoderWA} to obtain the binary document region mask. The object segmentation model is trained with a binary cross-entropy loss~\cite{PieterTjerkdeBoer2005ATO}.
    In addition, we employ an off-the-shelf OCR engine as a textline extractor to obtain text lines as supervision signals for text consistency. %For more details, see~\ref{Dewarp Regularizations}.
    
    \subsection{Dewarp Transformer}
    \label{section:3.2}
    Given an image of arbitrary resolution as input, the goal for document image dewarping is to predict pixel-wise displacements. In this paper, we introduce a module called Dewarp Transformer to complete the above task. The module consists of four sub-modules: Dewarp Head, Distorted Image Encoder, Dewarped Image Decoder, and Dewarp Tail. We downsample the preprocessed foreground image to obtain the model input $I_t\in \mathbb{R}^{H_t\times W_t \times C_t}$ before feeding it into the Dewarp Head, where $H_t = W_t$ is set to $432$ in our method to keep the text clear enough, and $C_t = 3$ is the number of RGB channels.

    \noindent
    \textbf{Dewarp Head.} The Dewarp Head is utilized for preliminary feature extraction, which consists of two convolutional layers and six residual blocks~\cite{KaimingHe2015DeepRL}. Dewarp Head extracts the image features from $I_t$ that is indicated as $z_v \in \mathbb{R}^{\frac{H_t}{8}\times \frac{W_t}{8} \times c_v}$, where $c_v = 384$ in our method. Then, we straighten the output features $z_v$ into a sequence of tokens $z_s \in \mathbb{R}^{N_v\times c_v}$ as input of the Distorted Image Encoder, where $N_v = \frac{H_t}{8}\times \frac{W_t}{8}$. 

    \noindent
    \textbf{Distored Image Encoder.} 
    In order to improve the global relationship between image pixels, we adopt the vanilla transformer~\cite {AshishVaswani2017AttentionIA} further to encode image features. 
    The transformer encoder incorporates the 2D position embedding from sinusoidal spatial position encoding for each token to append spatial information.
    Finally, the Distorted Image Encoder produces a visual representation $z_e \in \mathbb{R}^{N_v \times c_e}$ with global semantic information.

    \noindent
    \textbf{Dewarped Image Decoder.} We design the Dewarped Image Decoder with a six-layer transformer network~\cite{AshishVaswani2017AttentionIA} and feed the visual embedding $z_e$ and a learnable embedding $p_e \in \mathbb{R}^{N_d \times c_d}$ into the decoder to obtain predicted backward map $z_{d} \in \mathbb{R}^{N_d\times c_d}$. Each decoder layer consists of a cross-attention attention module and a feed-forward network. We add embedding $p_e$ to the query and key in each cross-attention calculation.

    \noindent
    \textbf{Dewarp Tail.} The Dewarp Tail module contains an update block and an upsampling layer. The update block is a two-layer convolutional network to transform the decoder output into a backward map $z_p$. Following the DocTr~\cite{HaoFeng2022DocTrDI}, we use an upsampling layer to resize $z_p$ to the original scale and get the final displacement mapping $z_{map} \in \mathbb{R}^{H_0 \times W_0 \times 2}$, where $H_0$ and $W_0$ are the original sizes of the input image, two channels denote horizontal and vertical displacement map.

    \subsection{Dewarp Regularizations}
    \label{Dewarp Regularizations}
    As shown in the bottom  part of Figure~\ref{fig:3}, the MataDoc architecture proposes three optimization constraints in the training process: backward map regression, text constraint, and margin constraint. The backward map regression is the most commonly used method in previous works. However, for arbitrary boundary document dewarping, only using backward map constraint might tend to poor performance. Therefore, we design the margin and text aware regularizations to enhance boundary perception and keep text lines straight further. 
    The final integrated loss function in MataDoc is defined as follows:

    \begin{equation}
    	\begin{split}		
    		{L}_{total} &= {L}_{B} + \alpha {L}_{M} + \beta {L}_{T}. \\
    	\end{split}\label{eq:1}
    \end{equation}
    where ${L}_{B}$ denotes the regression loss on backward map, $L_{T}$ represents the constraint for control points coordinates of textlines with $L_2$ loss, and $L_{M}$ denotes the margin constraint with $L_1$ loss. $\alpha$ and $\beta$ are the weights associated to ${L}_{M}$ and ${L}_{T}$, respectively. In the following, we present the formulation of the three loss terms.

    \noindent
    \textbf{Backward Map Regression.} In the backward map based document dewarping methods, regression constraint is defined as the $L_1$ distance between the predicted backward map $z_p$ and the ground truth $z_{gt}$.
    
    \begin{equation}
    	\begin{split}		
    		{L}_{B} &= ||{z}_{gt}-{z}_{p}||_1, \\
    	\end{split}
    \end{equation}

    \noindent
    \textbf{Text Consistency Constraint.} As common sense, rectified text lines should be horizontally straight. In other words, the vertical coordinates of the central points on the same text line are as close as possible. 
    % Based on the above considerations, we take advantage of the vital clues of text lines to design a text line based on the regular constraint. 
    The specific calculation for text consistency constraint is as follows:
    \begin{equation}
    	\begin{split}		
    		{L}_{T} &= \frac{1}{T\times K}\sum_{i=1}^{T} \sum_{k=1}^{K} {||{CPs_{p}(y)}_{(ik)}-\overline{CPs_{gt}(y)}_i||^{2}_{2}}, \\
    	\end{split}
    \end{equation}
    where $T$ and $K$ denote the number of text lines and the number of control points on the text line. Here, $CPs_{gt}$ are sampled every 8 pixels from each text line extracted by textline extractor, $\overline{CPs_{gt}(y)}_i$ denotes the average vertical ordinate of the $i$-th text line on the rectified image. To obtain the ${CPs_{p}}$, we adopt the predicted BM ${z}_{p}$ to map $CPs_{gt}$ back on the distorted image, then use the $FM_{gt}$ to map the distorted control points on the rectified image.
    
    \noindent
    \textbf{Margin Consistency Constraint.} For arbitrary boundary documents, margins can be explored to enhance boundary perception. Specifically, the rectified mask can be obtained from the distorted mask $DM_{gt}$ using dewarp operation with backward map. The margin pixels in rectified masks should be remarked as background pixels. We use $L_1$ loss to constrain the margin prediction and ground truth.
    \begin{equation}
    	\begin{split}		
    		{L}_{M} &= ||\mathbf{M}_{gt}-\mathbf{M}_{p}||_1, \\
    	\end{split}
    \end{equation}
    where $\mathbf{M}_{gt}$ is the ground truth of the background mask on the rectified image, and $\mathbf{M}_{p}$ is the binary map of the background on the rectified image obtained by grid sampling using the predict backward map, $\mathbf{M}_{gt}, \mathbf{M}_{p} \in [0, 1]$.

\section{Dataset}

    We evaluate our method MataDoc on the ArbDoc, WarpDoc~\cite{ChuhuiXue2022FourierDR}, DocUNet~\cite{KeMa2018DocUNetDI}, and DIR300~\cite{HaoFeng2022GeometricRL} benchmark datasets, which effectively demonstrates the generality of MataDoc. In the following, we present the three datasets respectively.

    \subsection{ArbDoc Benchmark}
    We propose ArbDoc, a new benchmark for evaluating document image dewarping methods. ArbDoc benchmark consists of 188 document images captured by a smartphone camera in different scenes (supermarket, office, etc.) with different illumination conditions. To cover more document image acquisition scene, we collect documents from academic papers~\cite{GuillaumeJaume2019FUNSDAD}, magazines~\cite{KeMa2018DocUNetDI}, receipts~\cite{HuangZheng2019ICDAR2019CO}, books, food packaging, etc., and put them into four types of boundary scenes including complete, overflow, corner-absence, and occlusion as illustrated in Figure~\ref{fig:1}. 
    The proportion of distorted images on ArbDoc is shown in Table~\ref{tab:1}. Since ArbDoc covers abundant scenarios including tickets (20\%), food labels (20\%), documents (20\%), and books (40\%).
    Moreover, we manually annotate OCR results in the ArbDoc benchmark for evaluating the OCR performance of dewarping methods more accurately. The ArbDoc benchmark is the first challenging benchmark that contains incomplete documents and OCR annotations.

    \begin{table*}[h]
        \begin{center}
        \caption{Statistics of the ArbDoc benchmark dataset.}
        \setlength{\tabcolsep}{1.5mm}{
        \renewcommand{\arraystretch}{1.2}
        % \resizebox{0.89\textwidth}{!}{
        \begin{tabular}{l|cccc||ccccc}
        \hline
            {Types} & {Complete}  & {Overflow}  & {Absence}  & {Occulusion} & {Scenarios} & {Ticket} & {Food Label} &{Document} & {Book} \\ \hline
            {Nums} & {41} & {107} & {20} &{20} & {Nums} & {40} & {40} & {40} & {68}\\ \hline
        \end{tabular}}\label{tab:1}
        \end{center}
    \end{table*}

    \subsection{Other Benchmark}
    \noindent
    \textbf{WarpDoc Benchmark.} The WarpDoc benchmark~\cite{ChuhuiXue2022FourierDR} consists of 1,020 camera images of documents collected from scientific papers, magazines, envelopes, etc., with six different types of deformations including perspective, folded, curved, random, rotated, and incomplete. In particular, the "incomplete" subset (contains 177 images) considers missing corners and occluded boundaries.

    \noindent
    \textbf{DocUNet Benchmark.} The DocUNet benchmark ~\cite{KeMa2018DocUNetDI} consists of 130 distorted document images taken in various indoor and outdoor scenes. The documents for the DocUNet benchmark are warped into various distortions (curving, crumples, etc.) and collected from different types, such as receipts, letters, magazines, and academic papers.

    \noindent
    \textbf{DIR300 Benckmark.} The DIR300 benchmark~\cite{HaoFeng2022GeometricRL} contains 300 real document photos taken in various background and illumination conditions which suffer from four distortions, random curving, random folds, crumples, and flat. Before warping documents, the ground truths are captured by aligning four point corners of the regular rectangular document to get a perfect dewarping. 

\begin{table*}[h]
    \begin{center}
    \caption{Comparison with state-of-the-art methods on ArbDoc benchmark dataset regarding OCR accuracy. "↑" indicates the higher the better, and "↓" indicates the lower the better. Bold and underline denote the best and the second-best results, respectively.}
    \setlength{\tabcolsep}{1.5mm}{
    \renewcommand{\arraystretch}{1.2}
    \begin{tabular}{l|cccccccccc}
    \hline
    Dataset & \multicolumn{10}{c}{ArbDoc}
    \\\hline
        \multirow{2}{*}{Type}  & \multicolumn{2}{c}{Complete} & \multicolumn{2}{c}{Overflow} & \multicolumn{2}{c}{Absence} & \multicolumn{2}{c}{Occulusion} & \multicolumn{2}{c}{Total} \\ \cline{2-11}
        {} & ED $\downarrow$ & CER $\downarrow$ & ED $\downarrow$ & CER $\downarrow$  & ED $\downarrow$ & CER $\downarrow$ & ED $\downarrow$ & CER $\downarrow$ & ED $\downarrow$ & CER $\downarrow$ \\ \hline
        %\\
        {DewarpNet~\cite{DasSagnik2019DewarpNetSD}}
        &\underline{139.24}&\underline{0.2231}& 212.31& 0.2509 & 303.65& 0.2284 & \underline{209.85}& \underline{0.1784} & \underline{205.83} & \underline{0.2347} \\ 
        {DocTr~\cite{HaoFeng2022DocTrDI}}
        &200.95&0.3478& 242.74& 0.2956& 619.80&0.5879&360.95&0.4286&286.32 & 0.3522\\
        {DDCP~\cite{GuoWangXie2021DocumentDW}}
        &  200.04&0.2732&\underline{197.65}&\underline{0.2412}&\underline{285.40}&\underline{0.1957}&216.75&0.2025&{209.54} & {0.2392}  \\
        {DocGeoNet~\cite{HaoFeng2022GeometricRL}}
        &199.65&0.3306&238.72 &0.2918&631.75&0.5973 &346.15 &0.3938& 283.44 & 0.3436 \\\hline
        {MataDoc}      
        & \textbf{125.60} & \textbf{0.1859} &\textbf{165.40} & \textbf{0.1683} & \textbf{260.65} & \textbf{0.1694} & \textbf{169.95} & \textbf{0.1427} & {\textbf{167.34}} & \textbf{0.1695}\\ 
        \hline
    \end{tabular}}\label{tab:2}
    \end{center}
\end{table*}

\begin{table}[h]
    \begin{center}
    \caption{Comparison with state-of-the-art methods on WarpDoc~\cite{ChuhuiXue2022FourierDR} benchmark dataset regarding OCR accuracy.}
    \setlength{\tabcolsep}{1.5mm}{
    \renewcommand{\arraystretch}{1.2}
    \begin{tabular}{l|cc}
    \hline
    Dataset & \multicolumn{2}{c}{WarpDoc}
    \\\hline
        \multirow{2}{*}{Type}  & \multicolumn{2}{c}{incomplete} \\ 
        \cline{2-3}
        {} & ED $\downarrow$ & CER $\downarrow$ \\ 
        \hline
        {DewarpNet~\cite{DasSagnik2019DewarpNetSD}} & 1167.59 & 0.3226 \\ 
        {DocTr~\cite{HaoFeng2022DocTrDI}} & \underline{833.92} & \underline{0.2324} \\
        {DDCP~\cite{GuoWangXie2021DocumentDW}} & 1107.10&0.3162  \\
        {DocGeoNet~\cite{HaoFeng2022GeometricRL}} & 898.41 & 0.2522 \\\hline
        {MataDoc} & \textbf{417.29} & \textbf{0.1338} \\ 
        \hline
    \end{tabular}}\label{tab:2}
    \end{center}
\end{table}

\section{Implementation}
\subsection{Evaluation Metrics}

\noindent
\textbf{MS-SSIM.} Structural similarity (SSIM)~\cite{ZhouWang2004ImageQA} is an evaluation method of image structural similarity. Multi-scale structural Similarity (MS-SSIM) computes the weighted sum of SSIMs across multiple scales using a Gaussian pyramid to eliminate the dependence on sampling density.

\noindent
\textbf{LD.} Local distortion (LD)~\cite{ShaodiYou2016MultiviewRO} measures the average local deformation of each pixel, which is calculated as the average $L_2$ distance of matched pixels between the ground truth image and rectified image. The matched pixels are computed by the SIFT Flow ~\cite{CeLiu2011SIFTFD}, defined as a displacement field that maps each pixel between two images.

\noindent
\textbf{AD.} Aligned Distortion (AD)~\cite{KeMa2022LearningFD} is a metric for document unwarping that overcomes the drawbacks of both MS-SSIM and LD. AD aligns the undistorted image with the ground truth image by unifying translation and scale and weights the dewarping errors based on the gradient magnitude of pixels in the ground truth image.

\noindent
\textbf{ED and CER.} Edit Distance (ED)~\cite{VILevenshtein1965BinaryCC} is defined as the minimum number of operations required to convert a string to a reference string, including delete(d), insert(s), and replace(r). Character Error Rate (CER)~\cite{AndrewCMorris2004FromWA} quantifies the similarity of two strings, which was calculated by ED. $CER=(d+s+r)/N$, where $N$ is the total character number in the reference string.

\subsection{Experimental Setup}
    We implement our MataDoc in the PaddlePaddle~\footnote{https://www.paddlepaddle.org.cn/en/} framework and train segmentation module and rectification module on the Doc3D dataset ~\cite{DasSagnik2019DewarpNetSD} and PublayNet~\cite{zhong2019publaynet} dataset .
    
\noindent
\textbf{Preprocessing.} During the training segmentation module, we directly resize images to 512$\times$512. In order to generalize documents with incomplete boundaries, we add RandomCrop augmentation to randomly crop $N_b$ boundaries of documents in the Doc3D dataset ~\cite{DasSagnik2019DewarpNetSD}, where $N_b \in [1, 4]$. We use SGD optimizer with a batch size of 12. The initial learning rate is set as $1\times 10^{-2}$ and reduced based on the PolynomialDecay policy. The module is trained for 200000 iterations on one NVIDIA Tesla P40 GPU. For the textline extractor, we adopt the public OCR engine PaddleOCR~\footnote{https://github.com/PaddlePaddle/PaddleOCR/tree/dygraph/ppocr} to acquire text bounding box. Then, we obtain text line control points evenly on the midline of the bounding box.
    
\noindent
\textbf{Rectification.} During training, we remove the background of distorted images and add the RandomCrop operation as in the segmentation module.
 We use AdamW optimizer~\cite{IlyaLoshchilov2017DecoupledWD} with a batch size of 16. The learning rate reaches the maximum $1\times 10^{-4}$ after 700 iterations and is reduced based on Linear Policy. Our Dewarp model is trained for 50 epochs on 2 NVIDIA A100 40GB GPUS. The hyperparameters $\alpha$ and $\beta$ in Equation.~(\ref{eq:1}) are set as $0.1$ and $0.5$, respectively.
    
Since the text content in Doc3D benchmark~\cite{DasSagnik2019DewarpNetSD} is ambiguous, we adopt PubLayNet dataset~\cite{zhong2019publaynet} as the training dataset so that textline extractor can extract text line control points. The images in PublayNet are firstly dewarped by using the forward mapping of the Doc3D dataset, such that the distortion modes are the same as the Doc3D dataset, but with different document content.

   \begin{table*}[h]
    \begin{center}
    \caption{Quantitative comparisons on DocUNet~\cite{KeMa2018DocUNetDI} benchmark dataset regarding image similarity, distortion metrics, aligned distortion, and OCR accuracy.}
    \label{table:1}
    \setlength{\tabcolsep}{4mm}{
    \renewcommand{\arraystretch}{1.2}
    \begin{tabular}{l|ccccc}
    \hline
    {Methods} &  MS-SSIM $\uparrow$  & LD$\downarrow$  & AD $\downarrow$ & ED $\downarrow$ & CER $\downarrow$   \\ 
    \hline
    \hline
    {DocUNet~\cite{KeMa2018DocUNetDI}}
    & 0.4094 & 14.4341 & 0.6900 & 2068.42/1371.51 & 0.4961/0.4300 \\
    
    {DewarpNet~\cite{DasSagnik2019DewarpNetSD}}
    & 0.4735 & 8.3909 & 0.3999 & {894.54/556.45} & {0.2401/0.2224} \\

    {FCN-based~\cite{GuoWangXie2020DewarpingDI}}
    & 0.4278 & 7.7706 & 0.4278 & 1624.08/982.32 & 0.3979/0.3212 \\
    
    {DocProj~\cite{GuoWangXie2021DocumentDW}}
    & 0.2928 & 18.1910 & 0.9582 & 1756.96/1212.93 & 0.4396/0.3994 \\
    
    {DDCP~\cite{GuoWangXie2021DocumentDW}}
    & 0.4726 & 8.9842 & 0.4235 & 1500.46/821.82 & 0.3775/0.2848\\

    {PaperEdge~\cite{KeMa2022LearningFD}}
    & 0.4700 & 8.4992 & 0.3945 & 825.48/\underline{398.95} & {0.2116/0.1630}\\
    {Marior~\cite{JiaxinZhang2022MariorMR}}
    & 0.4780 & \underline{7.4370} & 0.4033 & 823.80/649.22 & 0.2055/0.2289\\
    {RDGR~\cite{XiangweiJiang2022RevisitingDI}}
    & 0.4929 & 9.1080 & 0.4613 & 754.52/{434.20} & \underline{0.1775}/\underline{0.1599} \\

    {DocTr~\cite{HaoFeng2022DocTrDI}}
    & \underline{0.5105} & 7.7582 & \underline{0.3698} & {749.50/481.35} & {0.1893/0.1789} \\
    
    {DocGeoNet~\cite{HaoFeng2022GeometricRL}}
    & 0.5040 & 7.7128 & 0.3825 & {\underline{723.32}/437.08} & {0.1887/0.1659}  \\
    
    {FDRNet~\cite{ChuhuiXue2022FourierDR}}
    & \textbf{0.5420} & 8.2051 & 0.4035 &837.30/500.91 & 0.2112/0.1892 \\
    
    \hline
    \textbf{MataDoc}      
    &  0.5029  &  \textbf{7.4263}  & \textbf{0.3154}  &  {\textbf{646.12}/\textbf{392.48}} 
    &  {\textbf{0.1692}/\textbf{0.1505}} 
    \\ \hline
    \end{tabular}}\label{tab:3}
    \end{center}
    \end{table*}

\subsection{Experimental Results}
    We evaluate the performance of MataDoc on three benchmarks for document image dewarping tasks in terms of image similarity, aligned distortion, distortion metrics, and OCR accuracy. The comparative experiments of ArbDoc, WarpDoc~\cite{ChuhuiXue2022FourierDR}, DocUNet~\cite{KeMa2018DocUNetDI}, and DIR300~\cite{HaoFeng2022GeometricRL} benchmark datasets can be seen in Tables~\ref{tab:2},~\ref{tab:2}, ~\ref{tab:3}, and~\ref{tab:4}, respectively. Thank the authors of DocUNet, DewarpNet, FCN-based, DocTr, Marrior, PaperEdge, FDRNet, RDGR, and DocGeoNet, they send us the rectified document images of the DocUNet benchmark so that we can directly evaluate the results.
 
    On the proposed ArbDoc benchmark dataset, we compare MataDoc with existing methods that have official code and public pre-trained models, including DewarpNet, DocTr, DDCP, and DocGeoNet. Since we directly manually annotated the OCR results of the document, we only evaluated the OCR metrics including ED and CER, as shown in Table~\ref{tab:2}. We can clearly see a significant improvement in both metrics. 
    Existing methods require complete document boundaries, and the missing boundaries of documents will affect the correction results. This is mainly due to the fact that the incomplete boundary causes a change in the overall shape of the document, which affects the position estimation of the document in the image.
    It can be seen that MataDoc can remove various geometric and appearance distortions and achieve the best performance even from dewarped documents with incomplete boundaries. MataDoc achieves state-of-the-art dewarping performance at both complete and incomplete boundaries, demonstrating that the proposed MataDoc is more general and robust to document's boundary changes compared with existing methods. On the WarpDoc benchmark, we also compare MataDoc with the baseline methods the same as ArbDoc's. We conduct experiments on the "incomplete" subset of the WarpDoc, as shown in Table~\ref{tab:2}, further demonstrating the effectiveness of MataDoc in arbitrary boundary cases.

    \begin{table*}[h]
    \begin{center}
    \vspace{-1em}
    \caption{Comparison with state-of-the-art methods on DIR300~\cite{HaoFeng2022GeometricRL} benchmark dataset regarding image similarity, distortion metrics, aligned distortion, and OCR accuracy.}
    \setlength{\tabcolsep}{0.8mm}{
    \renewcommand{\arraystretch}{1.2}
    % \resizebox{0.89\textwidth}{!}{
    \begin{tabular}{l|ccccc}
    \hline
        {Methods} &  MS-SSIM $\uparrow$  & LD $\downarrow$  & AD $\downarrow$ & ED $\downarrow$ & CER $\downarrow$  \\
        \hline
        \hline
        {DewarpNet~\cite{DasSagnik2019DewarpNetSD}}
        &0.4921 & 13.9480 & 0.3308 & {1120.00} & {0.3623} \\
        {DocTr~\cite{HaoFeng2022DocTrDI}}
        & 0.6160 & 7.2057 & 0.2580 & {744.07} & {0.2298}  \\
    
        {DDCP~\cite{GuoWangXie2021DocumentDW}}
        & 0.5524 & 10.9814 & 0.3602 & 2277.63 & 0.5844  
        \\
        {DocGeoNet~\cite{HaoFeng2022GeometricRL}}
        & \underline{0.6380} & \underline{6.4035} & \underline{0.2437} & \underline{709.22} & \underline{0.2244}  \\
        \hline
        {MataDoc}      
        &  \textbf{0.6389} & \textbf{5.7593} & \textbf{0.1785}  &  {\textbf{482.90}} 
        &  \textbf{0.1785}              
    \\ \hline
    \end{tabular}}\label{tab:4}
    \end{center}
    \end{table*}

    On the DocUNet benchmark, we compare MataDoc with previous state-of-the-art methods. For DocProj and DDCP, we generate rectified document images based on the official code and the pre-trained model and then evaluate the results. 
    For OCR accuracy evaluation, following DewarpNet and DocTr, we also select 50 and 60 images from the DocUNet benchmark dataset.
    As shown in Table~\ref{tab:3}, our MataDoc achieves state-of-the-art performance on most metrics (LD, AD, ED, and CER) and outperforms all existing approaches by a large margin. 
    In order to fully balance the texture matrix MS-SSIM and LD, we introduce the Aligned Distortion (AD) metric as the final evaluation criterion of the texture metric. Compare with previous methods DocTr, and MataDoc achieve a relative improvement in AD by 5.44\%, and CER by 2.01\%/2.84\%, respectively. Similarly, it achieves a significant improvement in AD by 6.71\%, and CER by 1.95\%/1.54\%, compared to DocGeoNet, which had previously achieved the best performance. 
    Especially, FDRNet achieves very excellent MS-SSIM texture metric results because their method introduces a Fourier converter to guide the model to correct high-frequency information, while our model does not design an enhancement module. However, from the perspective of the LD and AD metric, the rectified results of our model are still better than that of FDRNet. Such results show the superior dewarping performance of MataDoc over the state-of-the-art methods. 

    \begin{table*}[h]
    \begin{center}
    \caption{Ablation experiments with different training datasets on DocUNet benchmark dataset.}
    {
    \setlength{\tabcolsep}{1.1mm}{
    \renewcommand{\arraystretch}{1.2}
    \begin{tabular}{l|ccccc}
    \hline
        {Methods} & MS-SSIM $\uparrow$  & LD $\downarrow$ & AD $\downarrow$ & ED $\downarrow$  & CER $\downarrow$  \\
        \hline
        \hline
        {base (Doc3D)} &  0.4876 & \textbf{7.4928} & 0.3338 & 432.73 & \textbf{0.1630} \\
        {base (PubLayNet)}
        & \textbf{0.4928} & 7.5073 & \textbf{0.3246} & \textbf{430.78} & {0.1697} \\
        \hline
    \end{tabular}}}\label{tab:5}
    \end{center}
    \end{table*}

    \begin{table*}[h]
    \setlength{\tabcolsep}{0.8mm}{
    \renewcommand{\arraystretch}{1.2}
    \begin{center}
    \caption{Ablation experiments based on regularization term constraints on DocUNet benchmark dataset.}
    {
    \begin{tabular}{cccc|ccccc}
    \hline
        id & bm & text & margin  & MS-SSIM $\uparrow$  & ED  $\downarrow$ & CER $\downarrow$ \\ 
        \hline
        \hline
        1 & {\checkmark} &  &  &   0.4841  &  892.06/519.40 & 0.2189/0.1826 \\
        2 & {\checkmark} &  & {\checkmark} &  0.5056  & 797.88/529.98 & 0.2015/0.1835 \\
        3 & {\checkmark} & {\checkmark} & & 0.5041  & 791.98/511.87 & 0.2041/0.1852 \\
        4 & {\checkmark} & {\checkmark} & {\checkmark} & \textbf{0.5065}  & \textbf{722.44/437.96} & \textbf{0.1797}/\textbf{0.1619} \\ 
        \hline
    \end{tabular}}\label{tab:6}
    \end{center}}
    \end{table*}

    On the DIR300 benchmark dataset, we compare MataDoc with methods that have publicly available code and publicly available models.  
    As shown in Table~\ref{tab:4}, the proposed MataDoc consistently outperforms all previous methods on dewarping documents with more complex backgrounds and various illumination conditions on all metrics. Note that the document images in the DIR300 dataset usually suffer much more complex distortion than document images in the DocUNet benchmark, indicating that MataDoc can better handle complex conditions by constraining text lines and margins.

    \begin{figure*}[h]
    \centering 
    \includegraphics[width=0.92\linewidth]{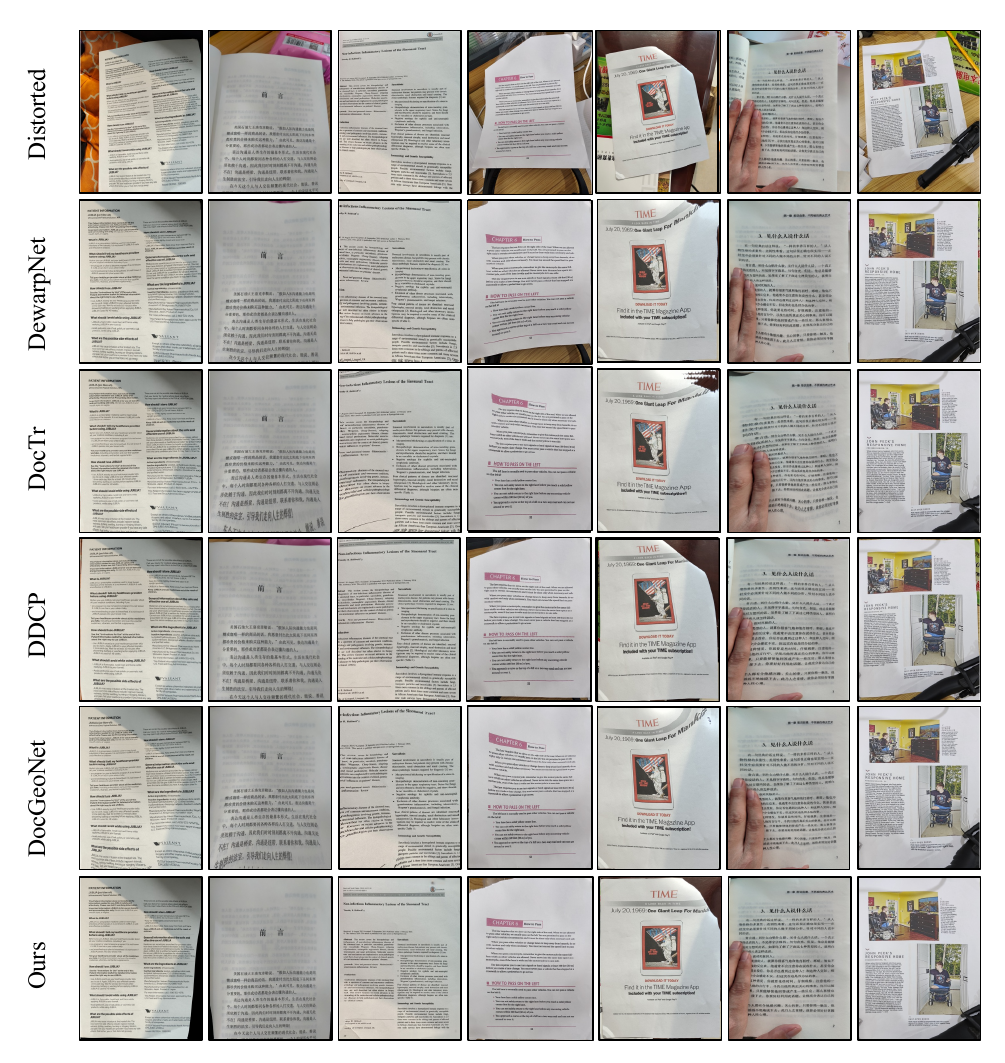}\vspace{-0.5em}
    \caption{Qualitative comparison with previous methods on the ArbDoc benchmark dataset. From top to bottom, Distorted, DewarpNet~\cite{DasSagnik2019DewarpNetSD}, DocTr~\cite{HaoFeng2022DocTrDI}, DDCP~\cite{GuoWangXie2021DocumentDW}, DocGeoNet~\cite{HaoFeng2022GeometricRL} and Ours.}
    \label{fig:4}\vspace{-0.5em}
    \end{figure*}
    
    In order to better demonstrate the effectiveness of our proposed MataDoc, we further perform a visual comparison with previous methods. Figure \ref{fig:4} shows a comparison of the qualitative performance of our proposed method on the benchmark ArbDoc dataset.
    For document images without boundaries, our text line consistency constraints play a major role. For example, consider the case with no boundaries (column 3), the other methods have a poor rectified effect in this case, while our rectified lines of text have a good flatness.
    For document images with boundaries, our constraints on boundaries and text lines complement each other. For example, if the boundaries of the document are incomplete (1, 2, 4, 5, 6, 7 columns), the result of other methods will be text line stretching at the boundaries. In addition, internal straight lines will also have an impact, and our text line and margin constraints can guide the model to rectify correctly.

\subsection{Ablation Studies}
    To investigate the relative contribution of each component in the proposed modules, we conduct a series of ablation experiments on DocUNet and ArbDoc benchmarks, which are illustrated in Table~\ref{tab:4} and Table~\ref{tab:5}, respectively. 
    
    \subsubsection{Training Dataset}
    As shown in Table~\ref{tab:5}, we evaluate the influence of the training dataset on the performance of MataDoc. We train the model on Doc3D and its variant using the PublayNet dataset as texture, respectively. 
    In the training dataset ablation study, data perturbations are added to both experiments.
    Note that for the OCR evaluation, we selected 60 images on the DocUNet dataset, following DocTr's setting.
    From the experimental results, we can see that the performance of MataDoc on PublayNet and Doc3D is relatively close. Due to the absence of illumination augmentation for the PublayNet dataset, text lines are clearer and more suitable for regularization. Hence, we chose PubLayNet as the texture image for the training set.
    
    \begin{table}[h]
    \begin{center}
    \caption{Ablation experiments based on regularization term constraints and data perturbation on ArbDoc benchmark dataset.}
    {
    \begin{tabular}{ccccc|cc}
    \hline
        id & bm & perturb & text & margin  & ED $\downarrow$ & CER $\downarrow$ \\ 
        \hline 
        \hline
        1 & {\checkmark} &  & & & 179.58 & 0.1890\\
        2 & {\checkmark} & {\checkmark} &  & & {172.76}&0.1790 \\
        3 & {\checkmark} & {\checkmark} & {\checkmark} &  & 169.74 & 0.1768\\
        4 & {\checkmark} & {\checkmark} &  & {\checkmark}  &   {167.75} &{0.1765} \\
        5 & {\checkmark} & {\checkmark} & {\checkmark} & {\checkmark} & \textbf{167.34} & \textbf{0.1695} \\ 
        \hline
    \end{tabular}}
    \label{tab:7}
    \vspace{-1em}
    \end{center}
    \end{table}

    \subsubsection{Dewarp Regularizations}
    To prove the dewarping capability of our method for document images on the arbitrary boundary, we remove some components to perform ablation experiments on the DocUNet and ArbDoc datasets. To verify the superiority of the proposed regularizations in complete boundary situations, the supervision of text and margin regularizations are gradually added to the baseline on DocUNet in Table~\ref{tab:6}. As we can see in experiments 2 and 3, the constraint on margins and text lines gains better performance, proving that regularizations can demonstrably guide the model for dewarping. Moreover, combining both the two regularizations can further improve dewarping performance, especially on OCR metrics, as shown in experiment 4. In addition, we perform an ablation study of data perturbation and regularizations in arbitrary situations. 
    As shown in Table ~\ref{tab:7}, data perturbation can bring a certain reduction in the error rate of OCR metrics including ED and CER, which indicates that data perturbation can enhance the robustness of the model. In comparison to experiments 2, 3, and 4, the text and margin constraints can bring ED reduction of 3.02 and 5.01, as well as 0.22\% and 0.25\% for CER, respectively, proving that both text line and margin constraints further significantly improve the performance of MataDoc. And the improvements brought by the text constraints and margin constraints alone are comparable. Moreover, combining the two constraints together can achieve a higher improvement since they can promote each other.
 
\section{Conclusion}
    In this work, we propose MataDoc, a novel document dewarping network based on margin and text aware regularization for the arbitrary boundary. We adopt margin regularization by considering background consistency to enhance boundary perception explicitly. Text aware regularization is introduced with word position consistency to keep text lines straight in unwarped images. In order to prove the effectiveness of MataDoc, we propose a novel benchmark ArbDoc, mainly focusing on incomplete boundary situations. MataDoc outperforms state-of-the-art methods on complete boundary benchmarks, meanwhile significantly improving dewarping performance on arbitrary boundary benchmarks, particularly in incomplete boundary situations. 

%%%%%%%%% REFERENCES
{\small
\bibliographystyle{ieee_fullname}
\bibliography{egbib}
}

\end{document}